\documentclass[letterpaper,10pt,conference]{ieeeconf}
\IEEEoverridecommandlockouts
\usepackage{amsmath,amssymb}
\usepackage{graphicx}
\usepackage{booktabs}
\usepackage{multirow}
\usepackage{array}
\usepackage{url}
\usepackage{microtype}
\usepackage{placeins}
\usepackage{balance}
\usepackage{stfloats}
\usepackage[style=ieee,backend=biber,sorting=none,giveninits=true,maxbibnames=3,minbibnames=1,doi=false,url=false,isbn=false]{biblatex}
\newcommand{\method}{C$^2$Nav}

\title{\LARGE \bf
C$^2$Nav: Compare Before You Commit for Zero-Shot Vision-and-Language Navigation
}

\author{
Runtian Zheng$^{1}$,
Congpeng Zhang$^{1,\dagger}$\thanks{$\dagger$Corresponding Author.},
Ying Liu$^{1}$
\thanks{$^{1}$Runtian Zheng, Congpeng Zhang and Ying Liu are with North China University of Technology, Beijing, China. Email: zcp@ncut.edu.cn}
}

\begin{document}
\maketitle

\begin{abstract}
Zero-shot vision-and-language navigation in continuous environments (VLN-CE) increasingly places foundation vision-language models (VLMs) inside the navigation loop. Existing systems commonly request cardinal outputs such as waypoints, pixels, headings, progress values, or absolute arrival decisions, coupling a generative response to geometric magnitude or an irreversible commitment. We study a complementary model--robot interface: the VLM compares controller-constructed alternatives, while geometry, thresholds, action magnitude, and execution remain on the physical side. We instantiate this idea in \method, a training-free framework with three coordinated faculties. \emph{Seeing} performs ordinal Gaze Election over physically vetted candidate views; \emph{Remembering} maintains a compact route sketch and compares adjacent instruction-leg hypotheses; and \emph{Arriving} combines a hesitation ladder, look-back comparison, and revocable walk-back for reliable stopping. On the public OpenNav\_R2R-CE\_100 protocol, \method{} with Qwen3-VL-8B-Instruct obtains 41.0\% OSR, 31.0\% SR, and 16.7\% SPL, while the same interface with the standard GPT-5.5 model reaches 54.0\% OSR, 44.0\% SR, and 29.0\% SPL. Whole-faculty ablations reduce SR to 14.0\% without Seeing, 25.0\% without Remembering, and 29.0\% without Arriving. Matched role inversions that replace only the comparative answer form with cardinal/absolute questions reduce SR to 12.0\%, 28.0\%, and 21.0\% in the spatial, transition, and terminal slots, respectively. The results indicate that a constrained decision interface and stronger VLM reasoning are complementary rather than interchangeable.
\end{abstract}

\section{INTRODUCTION}
Vision-and-Language Navigation (VLN) requires an embodied agent to ground a natural-language route description in visual observations and execute the implied motion in an unseen environment \cite{anderson2018r2r}. VLN in Continuous Environments (VLN-CE) removes the predefined navigation graph and exposes the agent to partial observability, collision avoidance, fine-grained motor control, and explicit stopping \cite{krantz2020vlnce}. The central difficulty is therefore not only understanding an instruction, but translating semantic judgment into reliable physical commitment over a long horizon.

Learning-based VLN has progressed through stronger cross-modal representations, history modeling, topological planning, and bird's-eye-view spatial reasoning \cite{hong2021vlnbert,chen2021hamt,chen2022dualscale,an2025etpnav,an2023bevbert,wang2023gridmm}. Foundation models enable a complementary zero-shot regime in which the navigation policy can be assembled from pretrained reasoning and perception without VLN-specific optimization. NavGPT and DiscussNav demonstrate explicit language reasoning and deliberation; MapGPT introduces map-guided prompting; Open-Nav studies zero-shot VLN-CE with foundation models; InstructNav and CA-Nav organize progress through value or constraint structures; and SmartWay augments waypoint navigation with history-aware reasoning and backtracking \cite{zhou2024navgpt,long2024discussnav,chen2024mapgpt,qiao2025opennav,long2025instructnav,chen2025canav,shi2025smartway}.

Recent systems make the model--controller boundary still more explicit. LaViRA translates a language-level direction into a grounded visual target and robot action, and Uni-LaViRA extends this translation across navigation tasks and embodiments \cite{ding2026lavira,ding2026unilavira}. P2DNav separates panoramic direction selection from down-view pixel grounding; LightZeroNav separates local action execution from route-stage transition for lightweight VLMs; and AgenticNav expands model agency through callable action, depth, and memory tools \cite{sheng2026p2dnav,luo2026lightzeronav,li2026agenticnav}. These systems establish interface design as a central issue, but their online semantic outputs can still be a direction, pixel, action, progress state, or tool call. Our complementary question is narrower and directly testable: what is gained when model authority is deliberately restricted to ordering controller-generated alternatives before physical commitment?

A recurring design choice in these systems is the \emph{answer type} requested from the model. A VLM may be asked for a coordinate, waypoint, heading, progress magnitude, confidence, or binary arrival verdict. Such outputs are convenient because they can be consumed directly by a controller, but they also couple an uncalibrated generative model to geometric magnitude or an irreversible decision. A small heading error can create a long deviation; an optimistic progress estimate can switch sub-goals too early; and a permissive arrival criterion can terminate an otherwise recoverable episode.

Human wayfinding suggests a different abstraction. Landmark and route knowledge can be useful before precise metric survey knowledge is available \cite{siegel1975,golledge1999}: a traveler can decide which corridor better matches an instruction without estimating its bearing, or recognize that the current scene is more compatible with the next route segment without reporting a completion percentage. Comparative judgment has a related measurement advantage. In a simple signal-detection model, an absolute response depends jointly on evidence sensitivity and a movable response criterion, whereas a forced comparison depends primarily on evidence ordering \cite{thurstone1927,green1966}. Pairwise foundation-model judgments can still exhibit presentation-order effects \cite{zheng2023llmjudge}; the point is not that comparison is unbiased, but that its authority can be bounded and audited.

\begin{figure}[t]
    \centering
    \includegraphics[width=\columnwidth]{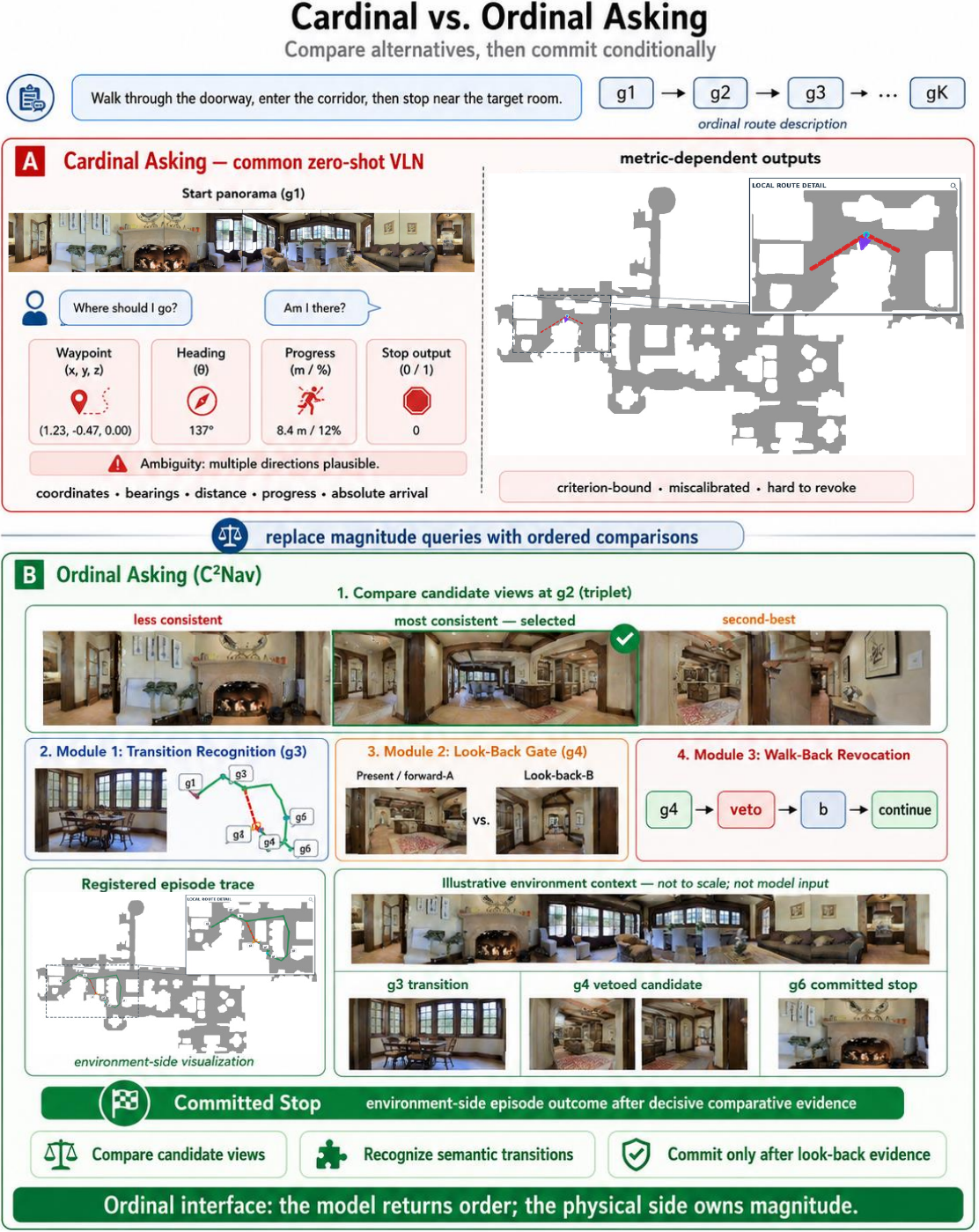}
    \caption{Cardinal asking versus the proposed ordinal interface. The VLM returns an ordering over controlled alternatives; metric state, bounded motion, and terminal commitment remain on the physical side.}
    \label{fig:cardinal-ordinal}
\end{figure}

We therefore formulate a \textbf{Compare-Then-Commit ordinal interface}. At decision-critical points, the VLM sees a finite set of alternatives and returns only a preference or ranking. Depth, odometry, traversability, thresholds, and motion magnitude are computed by deterministic modules. A short physical action is committed only after the semantic alternative has been selected, after which the agent re-observes and reasons again. This separation creates a natural experimental control: keep perception, history, and execution fixed, but replace the comparative question by a cardinal or absolute one in the same decision slot.

We instantiate the interface in \method{} through three faculties corresponding to three recurrent VLN-CE questions. \emph{Seeing} asks \emph{where should I move next?} and elects among physically valid views. \emph{Remembering} asks \emph{has the route entered the next instruction stage?} and compares neighboring route hypotheses against an episodic sketch. \emph{Arriving} asks \emph{is the current place sufficiently better supported than recently passed alternatives to stop?} and uses staged evidence, look-back comparison, and walk-back recovery before commitment. The design is hierarchical, but our emphasis is the message crossing the model--robot boundary rather than hierarchy alone.

Our contributions are threefold:
\begin{itemize}
    \setlength{\itemsep}{1pt}
    \setlength{\parsep}{0pt}
    \setlength{\topsep}{2pt}
    \item We introduce a practical \textbf{ordinal model--robot interface} that restricts VLM decisions to controlled comparisons while retaining geometry and execution on the physical side.
    \item We develop \textbf{\method}, a zero-shot navigation framework that integrates Gaze Election, route-sketch transition reasoning, and revocable terminal commitment into a unified closed loop.
    \item We evaluate the design with \textbf{whole-faculty ablations and matched role inversions}, showing that comparative decision form is especially important for spatial selection and terminal commitment, and remains complementary to backbone scaling.
\end{itemize}

\section{RELATED WORK}
\subsection{Vision-and-Language Navigation}
R2R established instruction-guided navigation in Matterport3D \cite{anderson2018r2r,chang2017matterport}. Early work improved generalization through instruction augmentation and cross-modal representation learning \cite{fried2018speaker,hong2021vlnbert}. HAMT models long observation--action histories explicitly, while DUET combines global graph reasoning with local action selection \cite{chen2021hamt,chen2022dualscale}. VLN-CE transfers the task to continuous Habitat environments \cite{savva2019habitat,krantz2020vlnce}, where the agent must execute low-level actions rather than jump between graph nodes. ETPNav, BEVBert, and GridMM further strengthen long-horizon planning with topological or map-like intermediate representations \cite{an2025etpnav,an2023bevbert,wang2023gridmm}. These methods establish strong continuous navigation baselines, but normally rely on task-specific learning or learned spatial modules.

\subsection{Foundation Models for Zero-Shot VLN-CE}
Foundation models make it possible to reason about navigation instructions without training a complete task-specific policy. NavGPT uses an LLM for explicit instruction-following reasoning, and DiscussNav decomposes navigation reasoning across multiple experts \cite{zhou2024navgpt,long2024discussnav}. MapGPT maintains a linguistic topological representation to support global planning \cite{chen2024mapgpt}. In continuous environments, Open-Nav combines foundation-model reasoning with executable waypoint candidates, SmartWay improves waypoint quality and history-aware backtracking, and InstructNav and CA-Nav organize progress through value maps or explicit constraints \cite{qiao2025opennav,shi2025smartway,long2025instructnav,chen2025canav}.

Waypoint-free systems are not interface-free. LaViRA and Uni-LaViRA use a language--vision--robot translation in which MLLMs choose a direction and then localize a visual target \cite{ding2026lavira,ding2026unilavira}. P2DNav similarly separates panoramic choice from pixel-level grounding, LightZeroNav structures action and transition reasoning around a lightweight Qwen3-VL backbone, and AgenticNav lets the VLM select target pixels, query metric depth, and recall visual memory through tools \cite{sheng2026p2dnav,luo2026lightzeronav,li2026agenticnav}. Concurrent systems extend this trend through direct three-view action prediction and future--past reasoning, graph-constrained instruction solving with backtracking, global--local--global trajectory auditing, and online spatial cognitive memory \cite{shi2025fastsmartway,yin2025gcvln,zheng2026threestep,deng2026spacevln}. The closest structural overlap is Three-Step Nav, whose local candidate selection and look-back audit resemble parts of Seeing and Arriving; AgenticNav is the closest recent contrast in interface philosophy because it broadens model agency, whereas \method{} bounds it. Our claim is therefore not a new three-stage organization. \method{} differs in two coupled respects: every online semantic slot accepts only an ordering over alternatives constructed and vetted by the controller, and matched role inversions isolate the requested answer form while holding perception, history, and execution fixed.

\subsection{Hierarchical Navigation, Memory, and Recovery}
Long-horizon navigation is commonly stabilized by decomposing global intent from local execution. In learning-based VLN, DUET separates global graph reasoning from local action selection, while ETPNav evolves a topological planning structure as the agent moves \cite{chen2022dualscale,an2025etpnav}. Zero-shot systems adopt analogous structure without training a complete navigation policy. InstructNav explicitly decomposes an instruction into a dynamic navigation chain, CA-Nav manages sub-instruction completion under constraints, and SmartWay uses historical context and backtracking to recover from poor waypoint decisions \cite{long2025instructnav,chen2025canav,shi2025smartway}. These results support two recurring lessons: progress should be represented explicitly rather than inferred from the current frame alone, and an incorrect local commitment should remain recoverable whenever possible. \method{} follows both principles, but uses a compact route sketch rather than a learned map and makes the transition/recovery decisions part of the same comparative interface used for spatial selection.

\subsection{Comparative and Sequential Decisions}
Comparative judgment formalizes decisions through relative evidence rather than an absolute scale \cite{thurstone1927,green1966}. Reject-option classification similarly motivates abstention when evidence is insufficient \cite{chow1970}, and sequential testing motivates delaying commitment until evidence accumulates \cite{wald1945}. Temporally extended options provide a natural abstraction for route segments \cite{sutton1999}, while change-point detection motivates comparing whether observations remain consistent with the current route leg or support a transition \cite{page1954}. These classical ideas are useful because embodied navigation contains several asymmetric decisions: a short local motion can be corrected on the next cycle, a stage transition changes the interpretation of future observations, and a stop action ends the episode. \method{} therefore applies stronger evidence requirements as commitment becomes harder to revoke. We use comparison, abstention, and accumulated evidence as engineering principles rather than as a cognitive model of VLMs.

\section{PROPOSED METHOD}
\label{sec:method}
\subsection{Problem Formulation and Overview}
An R2R-CE episode provides a natural-language instruction $I$, egocentric RGB-D observations, odometry, and a continuous action interface. Before execution, a language-only preprocessing step parses $I$ into an ordered route ladder $G=(g_1,\ldots,g_K)$, where each $g_k$ describes a semantic route leg without coordinates or metric subgoals. This one-time nonmetric representation does not directly trigger motion and is therefore outside the three online ordinal decision slots; our interface claim concerns model outputs that cross the model--robot boundary during execution. At step $t$, the agent maintains its physical state $s_t$, current leg index $\ell_t$, current observation $O_t$, and an episodic route sketch $\mathcal{R}_t$. The navigation policy can be written as
\begin{equation}
 a_t=\mathcal{C}\!\left(\mathcal{Q}_{\theta}(I,g_{\ell_t},O_t,\mathcal{R}_t),s_t\right),
 \label{eq:policy}
\end{equation}
where $\mathcal{Q}_{\theta}$ is a frozen VLM query and $\mathcal{C}$ is the deterministic physical controller. The defining restriction is that $\mathcal{Q}_{\theta}$ returns an ordinal decision; metric quantities required to realize $a_t$ are supplied by $\mathcal{C}$.

\begin{figure*}[t]
    \centering
    \includegraphics[width=0.97\textwidth]{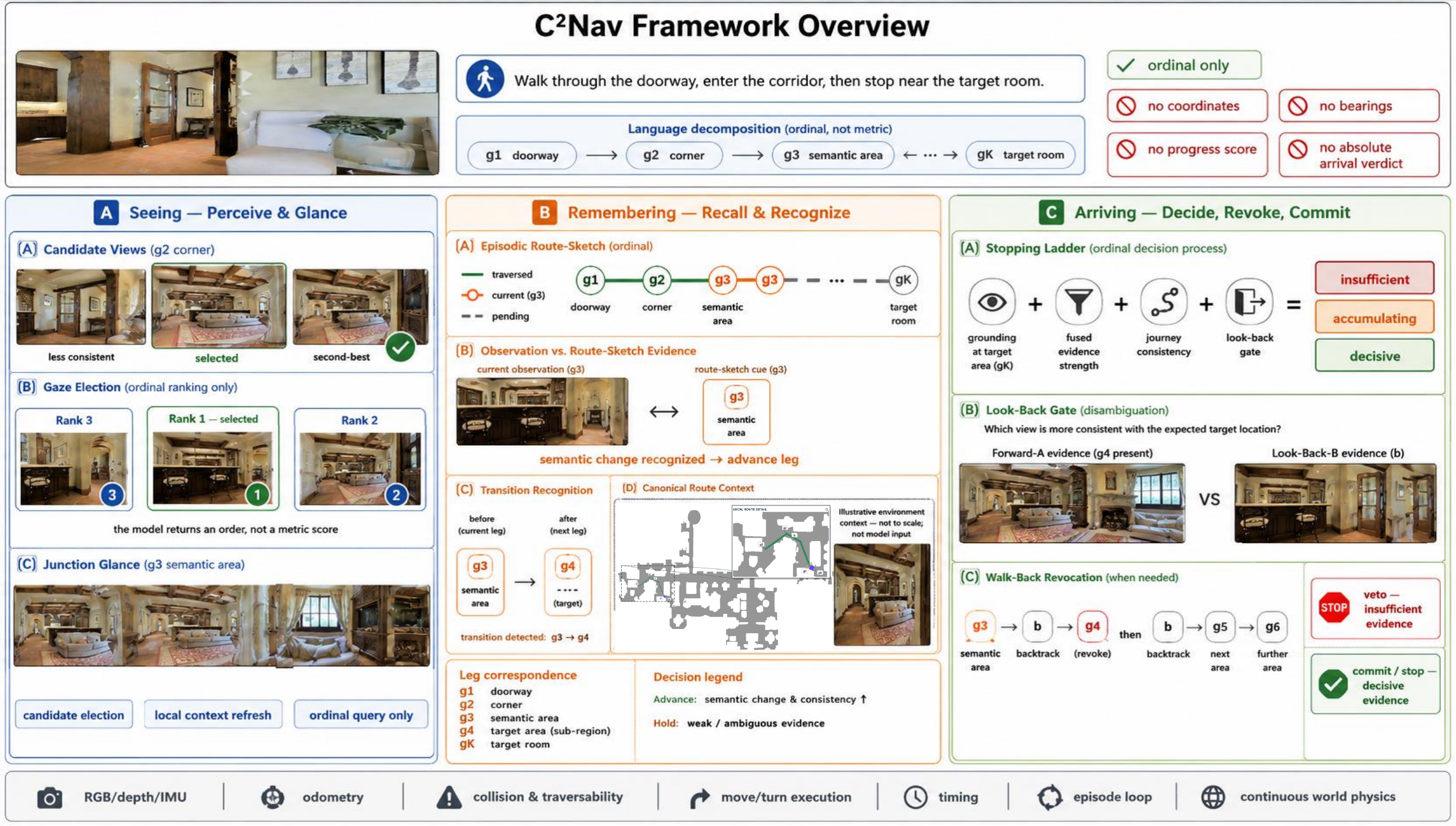}
    \caption{Overview of \method. \textbf{A Seeing} ranks physically vetted candidate views and can re-anchor through a junction glance. \textbf{B Remembering} maintains a compact route sketch and recognizes instruction-leg transitions by comparison. \textbf{C Arriving} accumulates terminal evidence, compares the present place with look-back evidence, and may revoke a rejected stop through Walk-Back. The top-down map is evaluation-side visualization rather than a VLM input.}
    \label{fig:framework}
\end{figure*}

The full architecture is shown in Fig.~\ref{fig:framework}. The three faculties operate at different temporal scales. Seeing is invoked frequently for short-horizon direction selection. Remembering accumulates route evidence and controls stage transitions. Arriving is activated only near the final instruction stage, where a false positive is particularly costly. This coarse-to-fine temporal organization lets the system spend stronger deliberation where commitment is harder to reverse.

\subsection{Ordinal Interface Contract}
For decision slot $j$, the controller constructs a finite alternative set $\mathcal{A}_j=\{a_j^1,\ldots,a_j^m\}$ and asks the VLM only for an ordering
\begin{equation}
 \pi_j=\operatorname{Rank}_{\theta}(I,O_t,\mathcal{R}_t;\mathcal{A}_j).
 \label{eq:rank}
\end{equation}
The VLM may return an index, ranking, or pairwise winner, but the interface does not request a pixel coordinate, heading, distance, progress percentage, scalar confidence, or direct motion duration. Once an alternative is selected, the physical side maps it to a bounded action using depth, odometry, collision checks, and fixed control limits. Thus, semantic preference and metric realization are separated.

A simple signal-detection surrogate motivates the restriction. For an absolute binary judgment with sensitivity $d'$ and response criterion $c$, the true-positive response under an equal-variance Gaussian observer is $\Phi(d'-c)$. For independent two-alternative evidence, the ordering probability is
\begin{equation}
 P_{\mathrm{2AFC}}=\Phi\!\left(\frac{d'}{\sqrt{2}}\right),
 \label{eq:2afc}
\end{equation}
which does not contain the free criterion $c$. We do not assume that VLM evidence is Gaussian or that prompting changes only criterion. Equation~\eqref{eq:2afc} merely motivates the controlled role-inversion experiments in Sec.~\ref{sec:experiments}, where the same semantic slot is queried in comparative and cardinal forms.

The contract is operational: every decision slot specifies what alternatives the VLM is allowed to compare and which quantities remain owned by the physical system. The compact ownership summary below makes this separation explicit. In particular, a model preference never bypasses feasibility checks, sets a metric threshold, or directly commits an irreversible stop. This shared ownership pattern is used across all three faculties rather than implemented as three unrelated prompt heuristics.

\begin{center}
\begin{minipage}{0.98\columnwidth}
\centering
\scriptsize
\textbf{Interface ownership at the three decision slots.}\\[2pt]
\setlength{\tabcolsep}{2.4pt}
\begin{tabular}{lll}
\toprule
Slot & VLM compares & Physical side retains \\
\midrule
Seeing & candidate views & free space; turn/step size \\
Transition & stay vs. next leg & displacement; switch timing \\
Arriving & current vs. look-back & stop gate; return path \\
\bottomrule
\end{tabular}
\end{minipage}
\end{center}

\subsection{Seeing: Gaze Election and Re-anchoring}
The physical side first renders a small local fan of candidate views (a triplet in the reported implementation)
\begin{equation}
 C_t=\{(v_t^i,o_t^i)\}_{i=1}^{m},
 \label{eq:candidates}
\end{equation}
where $v_t^i$ is an RGB view and $o_t^i$ is a controller-side traversability estimate obtained from depth and collision state. Candidates with insufficient free-space support are removed before the VLM is queried. Given the active route leg $g_{\ell_t}$, the VLM ranks the surviving views according to semantic consistency with the instruction. The body rotates toward the highest-ranked feasible view and executes only a short motion bundle before re-observing.

This bounded commitment is important in continuous navigation. A wrong semantic election affects only the current local rollout rather than defining a long open-loop trajectory. When the local fan is ambiguous or repeated local motion fails to expose useful evidence, a \emph{Junction Glance} constructs a wider panoramic strip from the sensor ring. The same ordinal query then ranks coarse sectors, the robot performs a pure re-orientation, and normal local Gaze Election resumes. Seeing therefore uses VLM semantics to choose \emph{which} controlled direction to favor while keeping \emph{how far} and \emph{how much to turn} outside the model.

\subsection{Remembering: Route Sketch and Contrastive Transition}
Long-horizon instructions require more than reactive view selection. Prompt-only histories can grow large and can lose physically meaningful progress after repeated summarization. We instead maintain an episodic route sketch
\begin{equation}
 \mathcal{R}_{t+1}=\mathcal{R}_t\oplus(\Delta \mathbf{x}_t,\tau_t,b_t,m_t,q_t),
 \label{eq:route}
\end{equation}
containing leg-level displacement $\Delta \mathbf{x}_t$, elapsed time $\tau_t$, coarse ordinal turning evidence $b_t$, a short recent visual/text trace $m_t$, and a compact journey-consistency cue $q_t$ indicating support for the current leg, the next leg, or ambiguity. The sketch is not a learned map or a model-authored metric state. Its role is to preserve controller-grounded evidence that the agent has actually traversed the route described by the instruction.

After a minimum physical-progress gate, the transition decision is phrased as a contrast between adjacent route hypotheses,
\begin{equation}
 H_{\mathrm{stay}}:g_{\ell_t}
 \quad \text{vs.}\quad
 H_{\mathrm{next}}:g_{\ell_t+1}.
 \label{eq:transition}
\end{equation}
The VLM chooses which hypothesis is better supported by the current scene and route sketch. A fixed controller-side consistency rule accepts a transition only after temporally repeated support, preventing a single noisy frame from advancing the route. This construction converts instruction progress from a scalar estimation problem into a local change-recognition problem: the system only needs to decide whether the evidence still belongs to the current leg or has become more compatible with the next one.

\subsection{Arriving: Hesitation, Look-Back, and Walk-Back}
Stopping receives a more conservative treatment because a premature terminal action cannot be corrected within the episode. Arriving therefore maintains a finite controller-side evidence state,
\begin{equation}
 \mathbf{e}_t=(e_{\mathrm{ground}},e_{\mathrm{journey}},e_{\mathrm{lookback}}),
 \quad e_r\in\{-1,0,+1\},
 \label{eq:ladder}
\end{equation}
where $r\in\{\mathrm{ground},\mathrm{journey},\mathrm{lookback}\}$, and $-1$, $0$, and $+1$ denote evidence against, undecided, and evidence for stopping. These codes are deterministic bookkeeping for categorical comparison outcomes; the VLM never emits a scalar confidence. A fixed evidence ladder maps $\mathbf{e}_t$ to \texttt{hold}, \texttt{inspect}, or \texttt{stop}, subject to physical vetoes such as implausible depth, insufficient route progress, or collision-induced inconsistency. The first level checks whether final-leg semantics are grounded in the current view; the second checks whether the journey recorded in $\mathcal{R}_t$ is compatible with the destination; and the final level spends a limited look-back budget only when the evidence remains ambiguous.

For look-back, odometry selects a previously visited frame $F^{-}$ at a fixed path distance behind the current frame $F_t$. The terminal query asks which frame is more consistent with the destination description. Presentation order can be swapped across repeated comparisons, and disagreement leads to abstention rather than forced stopping. A rejected stop candidate is not necessarily discarded. The controller can record a \emph{Walk-Back marker}; if subsequent motion produces disconfirming evidence, the robot returns to the candidate neighborhood and evaluates the terminal evidence again. This converts some conservative false negatives into recoverable decisions while maintaining a high bar for irreversible stopping.

\subsection{Closed-Loop Execution}
At each navigation cycle, \method{} (1) forms and vets candidate views, (2) performs Gaze Election and bounded motion, (3) updates the route sketch, (4) tests current-versus-next instruction-leg evidence after sufficient physical progress, and (5) activates the Arriving ladder only in the final leg. Standard spatial decisions therefore remain lightweight, while transition and terminal reasoning are event-gated. All model variants share the same sensing, route representation, physical gates, candidate construction, and controller; only the frozen VLM backbone or the queried answer type is changed in the experiments.

\textbf{Event schedule and fallback.} Seeing is the default per-cycle query because its consequence is deliberately short-horizon. Transition is queried only after the physical progress gate is satisfied, and Arriving is queried only in the terminal route leg. If the model returns an invalid ranking, all alternatives fail deterministic feasibility checks, or the evidence needed by a gated faculty is unavailable, the system does not synthesize a free-form action. It instead keeps the current route state, re-observes or re-anchors, and retries through the same bounded interface. This fallback keeps uncertainty from silently changing the action vocabulary.

\textbf{Shared instantiation.} The same frozen VLM can serve all three slots with slot-specific comparison templates. Candidate generation, depth handling, odometry, route-state updates, stopping thresholds, and low-level motion are independent of the VLM family. Consequently, backbone scaling changes semantic capacity without changing the controller, while the role-inversion experiments change the requested answer form without changing perception or execution. This separation is what makes the two experiment families directly interpretable.

\section{EXPERIMENTS}
\label{sec:experiments}
\subsection{Experimental Setup}
\textbf{Environment and dataset.} We evaluate in Habitat \cite{savva2019habitat} on R2R-CE validation-unseen \cite{krantz2020vlnce} using the public \texttt{OpenNav\_R2R-CE\_100} file released with Open-Nav \cite{qiao2025opennav}. LaViRA reports this standard 100-episode setting, and AgenticNav explicitly uses the exact same 100 episodes \cite{ding2026lavira,li2026agenticnav}. The protocol is used only for final evaluation, not prompt, threshold, or controller tuning; success requires stopping within 3~m. All variants share its episode IDs and initial states. Where baselines explicitly use this protocol, episode selection is aligned, although implementations, controllers, and model/API settings differ.

\textbf{Metrics.} We report Navigation Error (NE), Oracle Success Rate (OSR), Success Rate (SR), and Success weighted by Path Length (SPL). NE measures final distance to the goal; OSR measures whether the trajectory enters the success region at any point; SR requires correct final stopping; and SPL additionally penalizes inefficient trajectories. The tables report episode-level point estimates on the fixed subset. Matched ablations are interpreted directly, whereas small cross-paper differences are not treated as statistical significance.

\textbf{Implementation and controls.} The default checkpoint is \texttt{Qwen/Qwen3-VL-8B-Instruct}; 4B and 32B variants provide the scale comparison \cite{bai2025qwen3vl}. Proprietary runs use OpenAI model ID \texttt{gpt-5.5}, the standard model rather than GPT-5.5 Pro \cite{openai2026gpt55}. The VLM receives the instruction, stage-specific RGB candidates, and compact route context; depth and odometry remain controller-side. No VLN-specific fine-tuning, learned waypoint predictor, or task-specific policy optimization is used. All within-paper studies fix the route parser, sensors, candidate construction, physical gates, and controller: scale changes only the frozen VLM, and role inversion only the answer schema.

\subsection{Comparison with Reported Zero-Shot Methods}
Table~\ref{tab:main} reports the common 100-episode protocol, including LaViRA and Three-Step Nav as close hierarchical baselines and AgenticNav as the strongest recent interface-oriented contrast. With Qwen3-VL-8B-Instruct, \method{} reaches 31.0\% SR and 41.0\% OSR, matching InstructNav's SR but trailing LaViRA and Three-Step Nav overall. Standard GPT-5.5 raises \method{} to 44.0\% SR and 29.0\% SPL. AgenticNav broadens model authority through pixel-action, depth-query, and memory-recall tools and reports 55.0\% SR/48.41\% SPL with GPT-5.5 \cite{li2026agenticnav}. Thus we make no raw-SOTA claim; our complementary claim is the matched effect of restricting answer form under a fixed stack. Cross-paper values remain descriptive because controllers, prompts, implementations, and API settings differ. Full-split systems such as SpaceVLN remain broader context.

\begin{table*}[t]
\caption{Reported results on the common 100-episode R2R-CE validation-unseen protocol. Baseline values are transcribed from the cited primary papers; the LaViRA rows use its v2 manuscript \cite{ding2026lavira}. Episode selection is shared where the source reports the public protocol; implementation and API differences remain.}
\label{tab:main}
\centering
\scriptsize
\setlength{\tabcolsep}{3.1pt}
\begin{tabular}{lcccccc}
\toprule
Method & Backbone & WP-free & NE$\downarrow$ & OSR$\uparrow$ & SR$\uparrow$ & SPL$\uparrow$ \\
\midrule
Random & -- & -- & 8.63 & 12.0 & 2.0 & 1.5 \\
NavGPT-CE \cite{zhou2024navgpt} & GPT-4 & No & 8.37 & 26.9 & 16.3 & 10.2 \\
DiscussNav-CE \cite{long2024discussnav} & GPT-4 & No & 7.77 & 15.0 & 11.0 & 10.5 \\
MapGPT-CE \cite{chen2024mapgpt} & GPT-4o & Yes & 8.16 & 21.0 & 7.0 & 5.0 \\
Open-Nav \cite{qiao2025opennav} & GPT-4 & No & 6.70 & 23.0 & 19.0 & 16.1 \\
SmartWay \cite{shi2025smartway} & GPT-4o & No & 7.01 & 51.0 & 29.0 & 22.5 \\
InstructNav \cite{long2025instructnav} & GPT-4 & Yes & 6.89 & 47.0 & 31.0 & 24.0 \\
CA-Nav \cite{chen2025canav} & GPT-4 & Yes & 7.58 & 48.0 & 25.3 & 10.8 \\
Three-Step Nav \cite{zheng2026threestep} & GPT-5 (2025-08-07) & No & 5.87 & 39.0 & 34.0 & 29.12 \\
LaViRA \cite{ding2026lavira} & GPT-4o / Qwen2.5-VL-32B & Yes & $6.43{\pm}0.28$ & $43.3{\pm}3.2$ & $36.0{\pm}1.7$ & $28.3{\pm}0.8$ \\
LaViRA \cite{ding2026lavira} & Gemini-2.5-Pro / Qwen2.5-VL-32B & Yes & $6.54{\pm}0.27$ & $48.7{\pm}2.1$ & $38.3{\pm}0.6$ & $28.3{\pm}0.9$ \\
AgenticNav \cite{li2026agenticnav} & GPT-5.5 & Yes & 5.19 & 65.0 & 55.0 & 48.41 \\
\midrule
\method{} (Qwen3-VL-8B-Instruct) & Qwen3-VL-8B-Instruct & Yes & 8.89 & 41.0 & 31.0 & 16.7 \\
\method{} (GPT-5.5) & GPT-5.5 & Yes & 7.20 & 54.0 & 44.0 & 29.0 \\
\bottomrule
\end{tabular}
\end{table*}

\subsection{Backbone Scale Analysis}
A central question is whether the interface merely compensates for a weak model or remains useful as the backbone becomes stronger. Table~\ref{tab:scale} shows a consistent trend from Qwen3-VL-4B to 8B and 32B, with GPT-5.5 providing the strongest result among the tested backbones. SR increases from 28.0\% at 4B to 31.0\% at 8B, 39.0\% at 32B, and 44.0\% with GPT-5.5; SPL follows the same trend. These results indicate that the ordinal interface does not remove the need for model capability. Instead, stronger semantic reasoning and a constrained action interface contribute along complementary dimensions.

\begin{table}[t]
\caption{Backbone scale under the same \method{} interface, episode IDs, and physical controller.}
\label{tab:scale}
\centering
\scriptsize
\setlength{\tabcolsep}{3.8pt}
\begin{tabular}{lrrrr}
\toprule
Backbone & NE$\downarrow$ & OSR$\uparrow$ & SR$\uparrow$ & SPL$\uparrow$ \\
\midrule
Qwen3-VL-4B  & 9.30 & 36.0 & 28.0 & 15.5 \\
Qwen3-VL-8B-Instruct & 8.89 & 41.0 & 31.0 & 16.7 \\
Qwen3-VL-32B & 8.10 & 48.0 & 39.0 & 23.5 \\
GPT-5.5      & 7.20 & 54.0 & 44.0 & 29.0 \\
\bottomrule
\end{tabular}
\end{table}

\subsection{Ablation Studies}
We next isolate the three faculty-level innovations. Table~\ref{tab:faculty} reports both an additive construction and whole-faculty subtraction. Starting from the base zero-shot navigator, adding Seeing raises SR from 18.0\% to 25.0\%. Adding Remembering further raises SR to 29.0\% and substantially improves SPL, while the complete system reaches 31.0\% SR. This cumulative view shows that the gains are not produced by a single terminal heuristic.

The subtractive view is more diagnostic. Removing \textbf{Seeing} causes the largest degradation, from 31.0\% to 14.0\% SR, indicating that controlled semantic direction selection is the primary source of reliable local progress. Removing \textbf{Remembering} lowers SR to 25.0\% while OSR remains 41.0\%, showing that the agent can still approach the destination but is less reliable at managing stage progression and completion. Removing \textbf{Arriving} yields 29.0\% SR and 37.0\% OSR, supporting the value of terminal evidence accumulation and recovery even when a basic controller-side stopping path remains.

\begin{table*}[!t]
\caption{Ablation of the three \method{} faculties on the same 100 evaluation episodes. The upper block builds the full system cumulatively; the lower block disables one entire faculty while keeping the other two unchanged.}
\label{tab:faculty}
\centering
\scriptsize
\setlength{\tabcolsep}{4.6pt}
\begin{tabular}{lcccccc}
\toprule
Variant & Seeing & Remembering & Arriving & OSR$\uparrow$ & SR$\uparrow$ & SPL$\uparrow$ \\
\midrule
Base navigator & -- & -- & -- & 30.0 & 18.0 & 11.7 \\
+ Seeing (A) & \checkmark & -- & -- & 42.0 & 25.0 & 10.9 \\
+ A + Remembering (B) & \checkmark & \checkmark & -- & 37.0 & 29.0 & 16.2 \\
Full (A+B+C) & \checkmark & \checkmark & \checkmark & 41.0 & 31.0 & 16.7 \\
\midrule
$-$ Seeing (A) & -- & \checkmark & \checkmark & 30.0 & 14.0 & 6.1 \\
$-$ Remembering (B) & \checkmark & -- & \checkmark & 41.0 & 25.0 & 10.9 \\
$-$ Arriving (C) & \checkmark & \checkmark & -- & 37.0 & 29.0 & 16.2 \\
\bottomrule
\end{tabular}
\end{table*}

Component-level removals are consistent with the faculty analysis. Removing the Route Sketch gives 36.0\% OSR, 27.0\% SR, and 14.4\% SPL; removing the contrastive Transition gives 43.0\% OSR, 26.0\% SR, and 10.8\% SPL; and removing Walk-Back gives 36.0\% OSR, 26.0\% SR, and 14.1\% SPL. The Transition ablation is particularly informative: OSR increases while SR and SPL decrease, indicating that reaching the goal neighborhood is not sufficient when route-stage and stopping decisions are poorly coordinated.

\subsection{Role Inversion: Is Comparison Itself Important?}
Faculty ablation establishes that the modules are useful, but it does not isolate the proposed answer type. We therefore keep the same observation stream, route context, controller, and decision slot while replacing the comparative query with a cardinal/absolute counterpart. Table~\ref{tab:role} reports the resulting role inversions.

The spatial inversion has the largest effect: replacing ordinal Gaze Election with a direct commitment-style direction query reduces SR from 31.0\% to 12.0\% and SPL from 16.7\% to 5.3\%. The terminal inversion replaces comparative evidence accumulation with an absolute single-view arrival judgment and reduces SR to 21.0\%. The transition inversion is milder, reaching 28.0\% SR, which suggests that route memory and temporal consistency can partially stabilize stage switching even when the response form is less constrained. Across the three slots, the strongest sensitivity appears where an answer is immediately converted into physical direction or irreversible termination.

\begin{table}[!t]
\caption{Controlled role inversion under identical perception and execution.}
\label{tab:role}
\centering
\scriptsize
\setlength{\tabcolsep}{4.2pt}
\begin{tabular}{lrrr}
\toprule
Interface variant & OSR$\uparrow$ & SR$\uparrow$ & SPL$\uparrow$ \\
\midrule
Full ordinal interface & 41.0 & 31.0 & 16.7 \\
Spatial: cardinal commitment & 30.0 & 12.0 & 5.3 \\
Transition: cardinal progress & 40.0 & 28.0 & 14.8 \\
Terminal: absolute arrival & 28.0 & 21.0 & 9.7 \\
\bottomrule
\end{tabular}
\end{table}

\begin{figure*}[!b]
    \centering
    \includegraphics[width=0.94\textwidth]{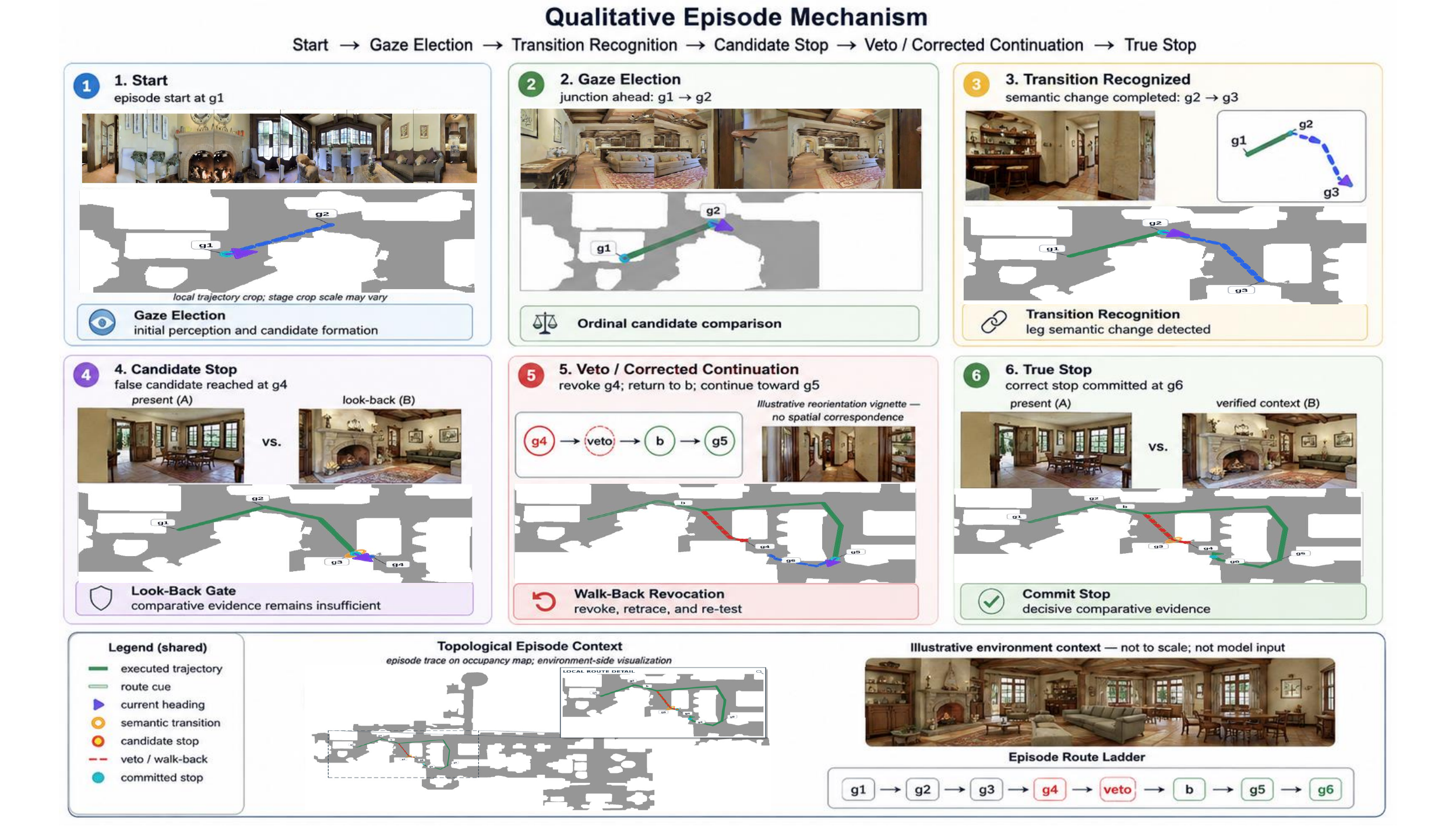}
    \caption{Qualitative \method{} trace. A plausible candidate stop can be withheld, navigation continues, and Walk-Back enables the agent to revisit and re-evaluate the terminal region before committing.}
    \label{fig:qualitative}
\end{figure*}

\subsection{Behavior Analysis}
The aggregate metrics reveal two distinct failure patterns. First, successful approach and successful stopping are not equivalent. The Full system reaches the success radius in 41 episodes and finishes successfully in 31, corresponding to a 75.6\% conversion from oracle success to final success. Removing Transition raises OSR to 43\% but lowers SR to 26\%, showing that trajectories can pass near the goal while stage and terminal decisions remain inconsistent. Second, recovery matters after a conservative decision. Without Walk-Back, SR and SPL decrease, consistent with losing the ability to revisit a previously plausible terminal region after subsequent evidence contradicts continued motion. These observations motivate treating Remembering and Arriving as coordination mechanisms rather than independent accuracy modules.

\subsection{Qualitative Analysis}
Figure~\ref{fig:qualitative} illustrates the intended closed-loop behavior. The agent first elects a locally plausible route direction, advances the instruction stage after the scene becomes more consistent with the next route leg, and then encounters an ambiguous terminal candidate. Instead of immediately stopping, the Arriving faculty compares the current scene against look-back evidence. When the evidence is insufficient, the stop is withheld and the agent continues. Subsequent observations can trigger Walk-Back, allowing the candidate region to be re-evaluated before final commitment. The mechanism is interpretable because each VLM judgment operates over a small, explicitly presented alternative set.

\section{DISCUSSION}
The experiments suggest that interface design and model scale solve different problems. Increasing backbone capacity improves semantic recognition and long-horizon reasoning, as reflected by the scale trend in Table~\ref{tab:scale}. The role inversions, however, show that a strong controller still benefits from limiting what form of answer is converted into action. This distinction is most visible in spatial and terminal decisions, where a cardinal error has immediate physical consequences. Transition decisions are less sensitive because they already aggregate route history and temporal evidence.

The ordinal restriction should not be interpreted as claiming that pairwise prompting is universally superior. Comparisons may still inherit order bias, visual hallucination, or correlated failures. Moreover, the route sketch is deliberately lightweight and does not provide the global planning capacity of a learned topological map. The present study is a controlled interface test on the public 100-episode protocol rather than the full validation-unseen split; the tables provide point estimates, and cross-paper implementations, controllers, simulator revisions, and API settings may differ. Latency, token cost, repeated-run uncertainty, and physical-robot validation are not characterized here. The proprietary-backbone result is also subject to model-version drift. Finally, the method assumes reliable RGB-D sensing and odometry; dynamic obstacles, long-term localization drift, and outdoor route semantics remain open challenges. These limitations bound the empirical claim without changing the matched evidence that answer form affects behavior under a fixed navigation stack.

\FloatBarrier
\section{CONCLUSION}
We presented \method, a zero-shot VLN-CE framework built around a simple design rule: compare, then commit. The VLM ranks controlled alternatives for spatial selection, route-stage transition, and terminal judgment, while the robot retains geometry, thresholds, and action magnitude. The resulting Seeing--Remembering--Arriving architecture achieves competitive zero-shot navigation and scales with stronger backbones. Whole-faculty ablations establish the complementary roles of the three components, while matched role inversions show that changing only the requested answer form can substantially alter navigation behavior. These results motivate treating the model--robot interface as a first-class design variable in foundation-model-driven embodied navigation.

\section*{Acknowledgment}
OpenAI GPT-5.6 Pro assisted before submission with language and organization edits in the Abstract, Introduction, Related Work, Method, and Discussion, plus reference checks. It did not generate experiments, data, code, or figures; the authors manually revised all wording and verified every technical claim, equation, result, and citation.

\balance
\printbibliography
\end{document}